# mini-ELSA: using Machine Learning to improve space efficiency in Edge Lightweight Searchable Attribute-based encryption for Industry 4.0

Jawhara Aljabri *†, Anna Lito Michala*, Jeremy Singer*, Ioannis Vourganas ‡
*School of Computing Science, University of Glasgow, United Kingdom
†Faculty of Computers and Information Technology, University of Tabuk, Saudi Arabia
‡School of Design and Informatics, Abertay University, United Kingdom

*Abstract*—In previous work a novel Edge Lightweight Searchable Attribute-based encryption (ELSA) method was proposed to support Industry 4.0 and specifically Industrial Internet of Things applications. In this paper, we aim to improve ELSA by minimising the lookup table size and summarising the data records by integrating Machine Learning (ML) methods suitable for execution at the edge. This integration will eliminate records of *unnecessary* data by evaluating added value to further processing. Thus, resulting in the minimization of both the lookup table size, the cloud storage and the network traffic taking full advantage of the edge architecture benefits. We demonstrate our mini-ELSA expanded method on a well-known power plant dataset. Our results demonstrate a reduction of storage requirements by 21% while improving execution time by 1.27x.

## I. INTRODUCTION

As emerging research in the Industry 4.0 domain is progressing, Industrial Internet of Things (IIoT) applications are proposed to improve the automation of factories and enable higher levels of control over the quality and quantity of produce [1]. The industry has traditionally collected data to support maintenance and operations [2]–[4]. However, with the constant collection of more diverse data new opportunities arise. Multiple stakeholders can request access to data or analysis of the data to ensure regulatory compliance, improve insurance premiums, or examine the quality of ordered products.

Such use cases however, demand that either the owner would analyse the data and provide the requested answer, or that the data is revealed to the requesting agent. Either approach introduces barriers. In the first case the owner must have the expertise to perform the analysis and must be trusted to provide the true results of the analysis [5]. In the second case issues relating to security, GDPR regulation or privacy of sensitive commercial information come to mind [6]. Continuous compliance has become a vivid new research domain as a result of these concerns. The continuous compliance solution for security compliance management is highly flexible and automated. This approach automates maintaining compliance with enterprise policies, regulations, and regulatory frameworks for managed IT services providers [7]. In earlier work [8], [9] Edge Lightweight Searchable Attribute-based encryption (ELSA) was proposed. ELSA allows multiple stakeholders to run their own queries (and hence analysis) over the data while providing cyber-security and privacy preservation by design. In this approach the process is automated, alleviating the burden of data analysis expertise and automatically complying with relevant regulation, provided that the owner has correctly described the access rules for each stakeholder. In this context continuous compliance can be verified by the regulator at any point in time without the direct involvement of the owner. This improves trust in the results of the analysis.

However, ELSA requires the use of intermediate storage on an edge server. This intermediate storage is analogous to the recorded data introducing scaling constraints [9]. To address this limitation in this paper we examine the combination of data preprocessing with our previously proposed method. Data preprocessing is well established in the domain of big data and can enable better performance for data analytics, reduces storage needs while extracting required meta-data for further processing [10].

Additionally preprocessing can address issues relating to the quality of the collected data. This is a fundamental challenge for IIoT applications where actionable information must be extracted from high quality data [11]. It is well known that data scientists invest 90% of a projects time in data preparation for ML or Artificial Intelligence (AI) [12]. This process is most often manual and provides results in non-real-time [11].

We propose the integration of an ML pipeline at the edge server to automate and address the aforementioned challenges and lead to the following contributions:

- reduction of the memory requirements on the edge server by 21% for ELSA through screening incoming data points.
- reduction of the volume and improvement of the quality of permanently stored data by evaluating the contribution of each data-point to the model's knowledge.

Through these contributions our methodology achieves the following benefits over and above the state of the art: (i) to minimise the ELSA system storage requirements; (ii) to minimise cloud storage costs; (iii) to optimise network traffic over the full stack; and (iv) to maintain high quality and optimal quantity datasets.

The remaining of this paper is structured as follows. Section II discusses related work in the domain of data pre-processing and cleaning. Section III introduces the proposed methodology. The evaluation method is presented in IV. The results of this evaluation are discussed in V. We conclude this work in VII.

## II. RELATED WORK

The researchers need to propose new or optimized pre-processing techniques for historical large datasets and data streams to provide dynamic pre-processing of big streaming data according to [10].

The data collected from different sources need to be processed for data quality, missing values and outlier detection [13]. Accelerating processing of data, which is unbounded, prevents any delay in further processing, as well as being important to making speedy and intelligent decisions [14].

The transformation of raw data, collected through different devices, is the prior requirement in data processing. The raw data has valuable and useful information and contains a large amount of noise, duplicate values, missing values, and inconsistency, depending upon the architecture. Thus, improving the raw data quality increases the efficiency and ease of data analysis. The process is also called data munging which commonly includes removing the unnecessary or invalid data which is not required for getting the underlying trends [15], [16].

The main task in data pre-processing is to eliminate noise and non-informative values and bring the reference parameters into standard form. Thus, Training the model on raw experimental data produces unexpected results [17].

Surveys, such as [18] conducted by Dogan et al., have shown that data selection is one of the main data preparation problems in Industry 4.0. A widespread challenge of ML application in manufacturing is selecting the data relevant to the analysis from the available database. All manufacturing data obtained from machine measurements may not always be used in the data mining process to solve the targeted problem. They can be related to different problems and become useless for that aspect. It is not apparent what part of the manufacturing dataset will be utilised at each point. This non-deterministic behaviour forces data miners to waste time in the non-beneficial data pile.

Factories leverage AI to transform information from various aspects of the manufacturing system into actionable insights. However, the data can contain a high degree of irrelevant and redundant information while the relevant part may be missing. These data curation issues present a challenge for the application of ML algorithms as the availability and quality of the manufacturing data strongly influence the performance and suitability of AI algorithms relative to expected results. Therefore, ensuring local data quality is the key to enabling a causal analysis of the manufacturing system [19]. One approach with promising results in automatically identifying anomalies in data is random isolation forest [20].

Finally in recent work [21] a new method was proposed to evaluate the contribution of each individual data point towards the output of a machine learning model. This approach is not ideal for cleaning the data from anomalies. However, it can be used to further evaluate which data points are useful for further analysis and which datapoints might not be needed in the future as the knowledge they bring has already been observed by the trained model. We believe this idea combined with anomaly detection can be used to automatically reduce the volume of data. This approach cah be used to automate the mining process separating out unecessary data points. We argue that this will improve the scalability of the previously propose ELSA method while improving automation for a variety of analytics performed over the collected data in Industry 4.0 applications that benefit from an edge architecture.

## III. METHODS

Our **mini-ELSA** method aims to combine two techniques used widely in the context of data pre-processing but often performed only on the training dataset manually and once at the begining of the data pipeline. We aim to screen data as they come in from IIoT sensors to maintain high quality and minimize the quantity. For this purpose we propose to combine an automated anomaly detection method and a well establised data valuation method in this new IIoT context. Our data pipeline methodology is presented in Fig. 1.

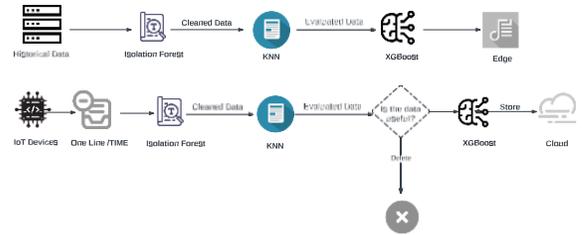

Fig. 1. mini-ELSA.

We start by dividing our dataset to a training set (90%) which in our IIoT context can be considered our *historical* dataset. At this stage our data is processed through our anomaly detection module which implements an Isolation Forest approach. We select this approach as our dataset has been proven to provide better performance with random forests according to [22]. In this paper the authors evaluated five machine learning algorithms, namely, K-Nearest Neighbors, Linear Regression, Gradient- boosted Regression Tree (GBRT), Artificial Neural Network, and Deep Neural Network of the Rapid Miner software suite. Keeping the default parameters, They evaluated the most crucial parameters of each algorithm to find the best of them that achieves minimum Root Mean Squared Error (RMSE) and Mean Absolute Error(AE). They also evaluated the effect of training set size and number of features on the achieved results. Consequently, GBRT outperforms the rest of the algorithms by achieving the least RMSE and AE with 450 trees while training on 90% of the dataset. Interestingly, it also exceeds in performance all the previously proposed methods on the same datasetby achieving the least RMSE and AE.

As a second step our *historical* data is valuated. For each datapoint a Shapley Value is calculated as a measure of the datapoint's contribution to the ML model that our exemplar industrial setup requires. This method of datapoint valuation was proposed by [21]. This approach is generalised for any ML application. However, it would require re-training when our mini-ELSA system is first deployed at a new setup or a new ML context. In this approach the value of a datapoint is estimated based on its potential contribution to the improvement of the model's predictive capacity. This approach uses the KNN method to implement efficient and performant Shapley Value calculations.

Once data is cleaned and valuated the data can be used to train any ML model. In this case we shall utilise the proposed GBRT approach from [22] and implement it via the xgboost library provided in python. This model will aim to predict the output of the industrial setup as a regression problem (e.g. power generated in a power station, or quality of product produced in a plastics factory). However, any appropriate model can be used depending on the use case.

Following the training phase the trained models (Isolation Forest, KNN, XGBoost) are deployed on the Edge server. Each new incoming datapoint arriving from any IIoT sensor can be screened in the same order for anomalies and valuation. If no anomaly and a high valuation are provided then the datapoint can be stored on the ELSA system, populating the Edge Server lookup tables and eventually the cloud server encrypted database.

*A. Implementation*

In this paper, we create an ML model pipeline and integrate it with the ELSA method presented in [8]. The ML pipeline was generated based on our methodology as presented in Algorithm 1.

In this model, we extract the following features: temperature (T), ambient pressure (AP), relative humidity (RH), and exhaust vacuum (V) and save them in the "X" object. Next, extract the parameter we want to predict (PE) and save it in the "Y" object. After that, we split the data into train and test subsets and then scaled to normalise them before running the Isolation Forest. Next, we create the Isolation Forest model with the number of isolation trees is 20, the number of samples is 50, and the contamination value is 0.1 and predicts the anomalies to remove, including outliers. Then we run the KNN regression and Shapley values calculation with testing and validation data as presented in [21]. Finally, we create an XGBoost model named "reg" with the best parameters: (the learning rate is 0.0075, and the number of the tree is 9000) and fit it with the training dataset.

---

**Algorithm 1** ML model

1: Load the data frame with data in a *df* variable
2: Extract the features (*AT, V, AP, RH*) columns and save them in the "*X*" object
3: Extract the paramater we want to predict (*PE*) and save it in the *Y* object
4: Split $X_{train} \leftarrow 90\%$, $X_{test} \leftarrow 10\%$
5: Split $Y_{train} \leftarrow 90\%$, $Y_{test} \leftarrow 10\%$
6: Scale $X_{train}$, $X_{test}$
7: **function** RANDOM ISOLATION($X_{train}$, *iforest*)
8: $\quad y_{pred} \leftarrow anomalies$
9: $\quad$ Remove $y_{pred}$ from $X_{train}, Y_{train}$
10: $\quad X_{trainiforest} \leftarrow X_{train}$
11: $\quad Y_{trainiforest} \leftarrow Y_{train}$
12: Create $x_{tst}, y_{tst}$ subsets
13: $(x_{tst}, y_{tst}) \leftarrow$ first 450 lines of $X_{test}, y_{test}$
14: Create $(x_{val}, y_{val})$ subsets
15: $(x_{val}, y_{val}) \leftarrow$ remaining of $X_{test}, y_{test}$
16: **function** GET TRUE KNN($X_{trainiforest}, x_{tst}$)
17: $\quad x_{tstknngt} \leftarrow results$
18: **function** GET TRUE KNN($X_{trainiforest}, x_{val}$)
19: $\quad x_{valknngt} \leftarrow results$
20: **function** UNWEIGHTED-KNNREGSHAPLEY(*TrainingData*)
21: $\quad x_{valspgt} \leftarrow results$
22: $\quad X_{trainKNN} \leftarrow X_{trainiforest}$
23: $\quad Y_{trainKNN} \leftarrow Y_{trainiforest}$
24: Create *idxs* object by sorting $g_{values}$
25: $keep_{idxs} \leftarrow idxs$
26: **function** XGB REGRESSION
27: $\quad$ Remove Shapley value from Training Data
28: $\quad$ **if** length $keep_{idxs} = X_{trainiforest}$ **then**
29: $\quad\quad X_{trainkeep} \leftarrow X_{trainiforest}$
30: $\quad\quad Y_{trainkeep} \leftarrow Y_{trainiforest}$
31: $\quad$ **else**
32: $\quad\quad keep_{idxs} \leftarrow X_{trainiforest}$
33: $\quad\quad keep_{idxs} \leftarrow Y_{trainiforest}$
34: Create xgboost name REG model
35: Train REG on ($x_{trainkeep}, Y_{trainkeep}$)
36: Fit REG on the best ($X_{trainKNN}, Y_{trainKNN}$)
37: Perform 5-fold cross-validation
38: Calculate MSE, RMSE, AE

---

## IV. EVALUATION METHOD

The goal of our methodology is to provide the following benefits over and above the state of the art solution, i.e., ELSA: (i) to minimise the edge storage requirements; (ii) to minimise cloud storage costs and network traffic; (iii) to achieve a higher performance; and (iv) without sacrificing the search or ML accuracy.

To evaluate the contributions presented at Section I we start by measuring the space requirements on edge and cloud for mini-ELSA versus the original ELSA implementation presented in [8], [9]. Further we present the accuracy of prediction on all three models to evaluate the effect of volume reduction towards predictive performance. We compare the predictive accuracy of the previously reported GBRT method in [22], the XGBoost approach without anomaly detection and data valuation, and our proposed XGBoost approach with the mini-ELSA implementation. In this work we do not evaluate the security and privacy preservation benefits of ELSA but we further present execution times to evaluate our proposed mini-ELSA overheads over and above the original ELSA method.

*A. Dataset and Use Case*

We use the well known Combined Cycle Power Plant (CCPP) dataset from the UCI machine learning repository [23]. The dataset contains 9,568 data points collected over 6 years (2006-2011). Features consist of hourly average ambient variables Temperature (T), Ambient Pressure (AP), Relative Humidity (RH) and Exhaust Vacuum (V) to predict the net hourly electrical energy output (EP) of the plant. Features consist of hourly average ambient variables as shown in Table I.

TABLE I
FEATURE DESCRIPTIONS OF CCPP DATASET SOURCED FROM [22]

| Features | Min | Max | Variance | Std |
|---|---|---|---|---|
| Temperature | 1.81°C | 37.11 | 55.54 | 7.45 |
| Ambient Pressure | 992.89 milibar | 1033.30 | 35.27 | 5.93 |
| Relative Humidity | 25.56% | 100.16 | 213.17 | 14.6 |
| Exhaust Vacuum | 25.36 cm Hg | 81.56 | 161.49 | 12.70 |
| Power | 420.26 MW | 495.76 | 291.28 | 17.06 |

*B. Experimental Setup*

We run the ML model and client application of ELSA on an edge device with an Intel 2.3 GHz Core i9 processor and 16GB RAM for the experimental setup. In addition, the server code was deployed on a docker container hosted on a DigitalOcean cloud provider in the UK. The plan for the cloud provider was CPU-Optimised, with one dedicated CPU, 2-32 vCPUs, 2TB Bandwidth, 2GB RAM/CPU and 50 GB backing storage. The results are calculated by taking the average of 100 runs for the two schemes, namely the original ELSA and mini-ELSA.

*1) XGBoost model:* To further confirm the reliability, robustness and validity of the developed model, the 5-fold cross validation method of 90% of randomly selected data was used to develop the model, and the remaining 10% of the data was used for model testing. Cross-validation is widely used for model selection, where the training set is repeatedly split into training and validation sets, where each split of the training set is used for training and at least one time for validation [24]. This process was repeated five times until each fold of the split data was tested once. The following Table II showed the average performance for all of the 5 folds.

For the ML model evaluation, we report three metrics appropriate for regression problems:
1) Mean Squared Error (MSE): it reports on the relationship between predictions and predicted values. It can

TABLE II
5-CROSS VALIDATION AVERAGE PERFORMANCE. THE RESULTS SHOW VARIATION IN RMSE PERFORMANCE AND AE.

| MSE | RMSE | AE |
|---|---|---|
| 12.38 | 3.51 | 2.47 |

highlight significant prediction errors and punish models which do not predict well. A perfect MSE is 0
2) Root Mean Squared Error (RMSE): it helps in presenting the real predicted value and not the squared. The perfect value is 0.
3) Mean Absolute error (AE): can show a linear relationship between the AE and predicted value.

In addition, we present the residuals plots which can expose any bias that the model might have. The residual, by definition, is equal to: *Residual = Observed − Predicted* value. Moreover, the residuals are sum to zero in simple linear regression, and they have a mean of zero positive values for residuals (on the y-axis). The plot shows if the prediction was too low (negative) or too high (positive), where 0 means the prediction was correct.

*2) mini-ELSA:* We have to categorise the data as our system is a keyword-based search. The data is divided into equal sizes based on the dataset's minimum and maximum value of the EP. We classify the predicted EP in this dataset into four classifications based on the value of EP: low, normal, high, and severe.
1) if the value of EP : between 420.26 and 439 then low
2) else if between 439 and 458 then normal
3) else if between 458 and 477 then high
4) else if between 477 and 495.76 then severe

For evaluation purposes, we considered the low,normal,high and severe values as the keywords.

To further evaluate the performance of our approach, we measure the search time using various keywords. The searched keyword are low, normal, high, severe. Also, we evaluate the overall execution time for original ELSA and proposed mini-ELSA implementation. In addition, we measure the lookup table size at the edge.

V. RESULTS

*1) Edge storage:* ELSA deploys a lookup table at the edge server to improve its search performance. The lookup table contains keywords that the query builder uses to construct optimal cloud search queries. To address goal I, mini-ELSA minimises the edge server storage by minimising the lookup table size (discussed in §III).

In Fig. 2, we measure and compare the lookup table size for ELSA and mini-ELSA using the CCPP dataset. In ELSA, the number of keywords stored in the lookup table is 9,568 while in mini-ELSA it is 7,624. We also measure the storage size of the lookup table. The lookup table in ELSA is 83.6 KB, while in mini-ELSA is 66.04 KB. Therefore, mini-ELSA minimses the edge storage by 21%. This result is the average of 100 repetitions and it is a reliable result as demonstrated by the error bars in Fig. 2.

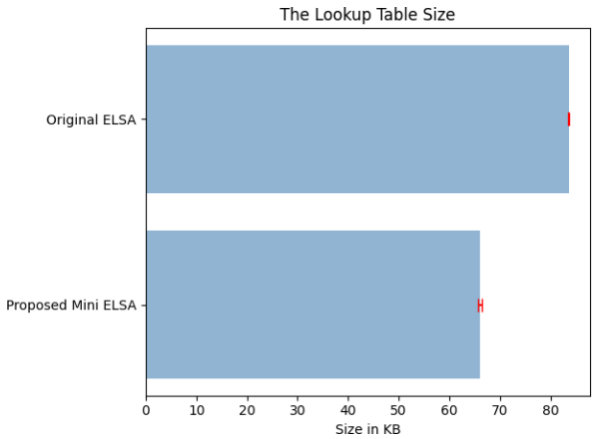

Fig. 2. Lookup Table Size averaged over 100 repetitions.

*2) Cloud storage:* To address goal II, we measure the ciphertext size on the cloud. For original ELSA it is 229.37 KB while for mini- ELSA it is 201.85 KB. Therfore, our mini-ELSA reduced the cloud storage by 12% again sustained over 100 repetitions.

*3) Overall Execution:* To address goal III we examine the overall performance of ELSA and mini-ELSA. Fig. 3 compares our mini-ELSA scheme with the original ELSA scheme in overall execution time. This graph aims to illustrate the overall effect of search time and lookup table size improvements to the latency experienced by the data user from initial query to final result. As demonstrated, the Mini ELSA scheme reduced the overall execution time by 21.98%.

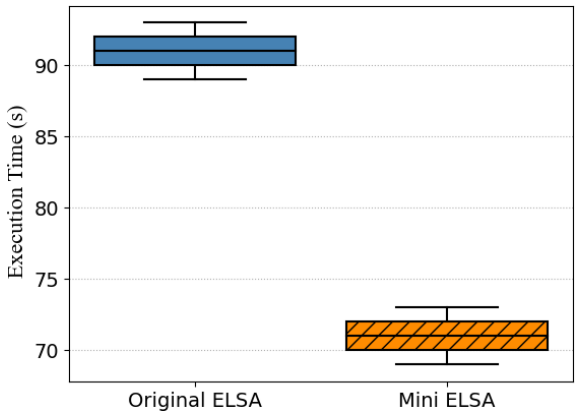

Fig. 3. Overall Execution Time in Seconds measured over 100 repetitions

*4) Search time:* To further evaluate the search performance of the mini-ELSA approach (as part of goal IV), we compare the search time for mini-ELSA with the original ELSA approach for the CCPP dataset using the four attributes (Fig. 4).

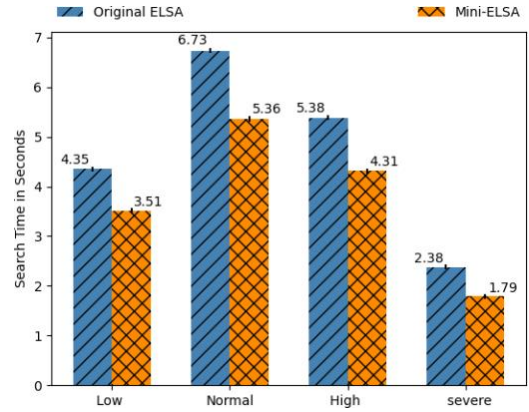

Fig. 4. Search Time in Seconds measured over 100 repetitions.

mini-ELSA achieves better performance than the original ELSA approach by 18.5% on average across all four cases.

*5) Prediction Accuracy:* To address goal IV in respect to predictive accuracy of the ML method, we examine the performance of the XGBoost model after the automated cleaning process using the anomaly detection and Shapley value calculation of mini-ELSA. We presented the observed Vs predicted PE values in Fig. 5 to understand the model's performance and linearity. The concentration across the diagonal demonstrates good linearity throughout the range of possible PE values supporting the suitability of the XGBoost regressor. It is evident that the model has high accuracy as there is strong correlation between the actual observed PE values of the test subset and the predicted PE values of the model.

We also examine the performance of the isolation forest after removing the outliers to establish its ability to correctly remove data points that deviate more than 3 standard deviations from the mean of each feature assuming normal distribution. The accuracy of this algorithm reach to 91% acco to our experiment.

Further, the residuals plot is presented in Fig. 6. The variance of the residual is not increasing with the predicted values. Also, there is no systematic curvature in the residual. Hence, there is a linear relationship and the regression approach is suitable for our problem. As a result, we can assume that the error is normally distributed, homoscedastic, and independent. This demonstrates the absence of bias in our approach improving trust in the predictive output.

However, the residuals plot demonstrates the presence of outliers. Those are points that higher that +10 or lower than -10 on Fig. 6. Even though they represent a very small minority, these outliers can influence the model's fit towards edge cases.

To investigate further the presense of predictive outliers, we plot a Q-Q residual plot (standardised VS theoretical quantiles) in Fig. 7. This graph observes the skew of the regression as an indication of a feature biasing the predictive performance. We observe the presence of the outliers in both ends of the Q-Q graph. However the majority of the points reside on the diagonal again demonstrating that the error follows a normal and equal distribution again supporting the absence of bias in the prediction.

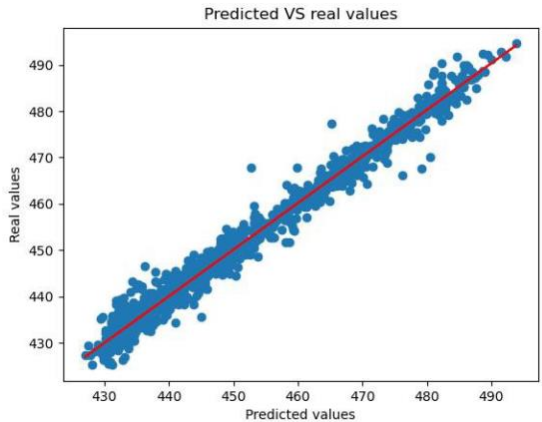

Fig. 5. Predicted vs Real values of the XGBoost model used for the power plant decision support at the end of the mini-ELSA ML pipeline.

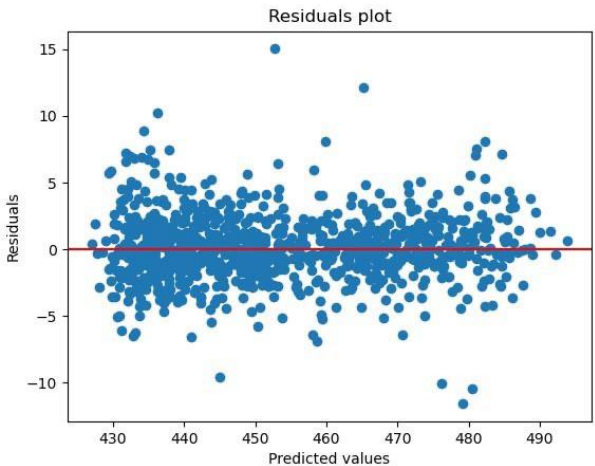

Fig. 6. Residuals plot for XGBoost model with mini-ELSA.

TABLE III
RMSE AND AE.

|  | RMSE | AE |
|---|---|---|
| Previous literature GBRT method in [22] | 2.583 | 1.856 |
| Original XGBoost approach | 2.581 | 1.942 |
| XGBoost with the mini-ELSA | 2.533 | 1.851 |

Finally, the effect of the most crucial parameters of the respective algorithms on the predicted power is presented in terms of both RMSE and AE. For the comparison, the dataset is split randomly into 90-10. we compare our XGBoost approach with the mini-ELSA implementation with XGBoost approach without anomaly detection and data and with Previous literature GBRT method in [22] in Table III. As we observe from the table the automated dataset cleaning proposed in mini-ELSA improves the overall predictive performance of the most suitable regressor as presented in [22]. At the same time the correlation of determination ($R^2$) for the predicted versus real PE values of the mini-ElSA approach was 0.977 for the test dataset. This in combination with the low RMSE supports the robustness and improved performance of mini-ELSA over and above the state of the art regressor applied on this power plant dataset. Thus, mini-ELSA not only automates a manual process but further improves trust in the predictive performance of the power plant's decission support system.

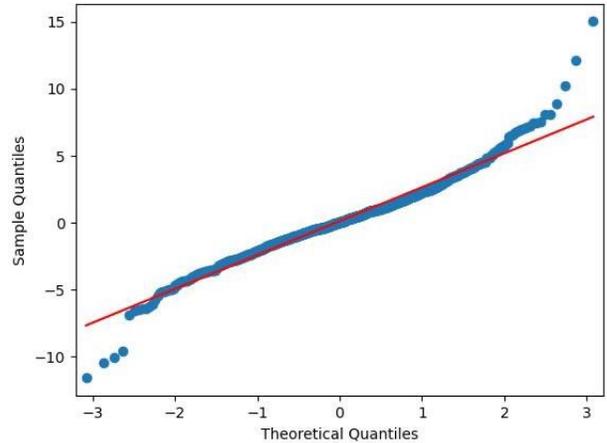

Fig. 7. Q-Q plot of the XGBoost model with mini-ELSA.

*6) Trade-off between data storage and accuracy:* It seems that there is a trade-off between data storage and accuracy. We have characterized one point in this trade-off space this was the minimum data storage for which accuracy is comparable with the original algorithm. Thus, we have run a mini-ELSA model with a different fraction of the whole data available after isolation forest and KNN such as 10%, 20% as presented in Fig. 8. Moreover, we collected the difference from the best RMSE value for each case. we observe a decline to accuracy with larger removals of data however it might be an acceptable decline for some use cases.

## VI. DISCUSSION

Our proposed mini-ELSA methodology can retrain and parameterise to be suitable for any edge IIoT solution. Therefore, we do not need to change the framework or ML pipeline for mini-ELSA. However, to apply a new use case (new dataset), the parameters in the k-mean, isolation foreset, KNN and xgboost in mini-ELSA need to be appropriately tuned to the best hyperparameters.

The CCPP dataset represents the sensor data from an industrial setting (it is related to the industrial use case because of our proposed mini-ELSA for industrial applications), and it is an excellent example dataset which is used in the literature and characterise as a well defined dataset [25]–[27].

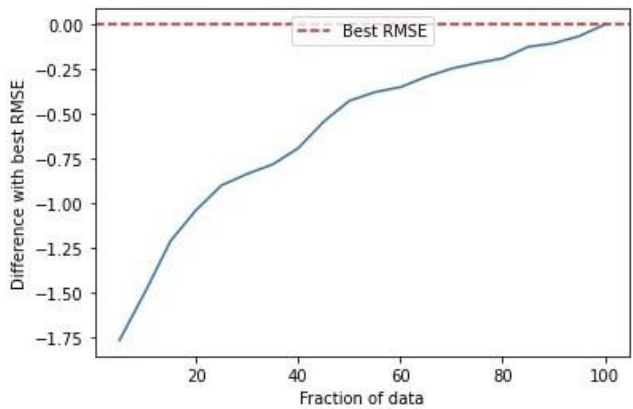

Fig. 8. Size vs RMSE trade-off curve.

For the limitation of the Isolation Forest, removing more outliers will increase RMSE in the final step. Also, in this dataset, the mean of PE (predicted value ) has a mean of $4\tilde{\,}00$. So a difference of 0.5 is minor considering the data; it corresponds to 0.13

## VII. CONCLUSION

In this paper, we integrate an ML pipeline with the previously published ELSA method that supports a cloud-edge architecture for Attribute-based Searchable Encryption with an optimised query process. The results show that our mini-ELSA reduced the original ELSA method Lookup table size at the Edge by 21% and consequently reduced the cloud storage by 12%. Further, the mini-ELSA approach improved the execution time of ELSA by 1.27x providing performance benefits. Finally, the mini-ELSA pipeline does not only reduce the volume of data it improves predictive performance by 1.11 % while automating the data mining process.

In future work, we aim to demonstrate the mini-ELSA methodology on a plastics factory use case to demonstrate transferability and evaluate the method's generalisation. Further, we aim to investigate the limitations of the proposed methodology, such as the hyperparameter tuning of the utilised models for each use case. Moreover, XGboost could also be configured to incrementally retrain on valuable new datapoints based on their Shapley values.